\documentclass{article}
\usepackage{amsmath,epsfig}
\usepackage{hyperref}
\usepackage[preprint]{spconfa4}
\usepackage{caption}
\let\OLDthebibliography\thebibliography
\renewcommand\thebibliography[1]{
  \OLDthebibliography{#1}
  \setlength{\parskip}{0pt}
  \setlength{\itemsep}{0pt plus 0.3ex}
}

\begin{document}\sloppy

\def\x{{\mathbf x}}
\def\L{{\cal L}}

\title{Expressive Speech-driven Facial Animation with\\ controllable emotions }

\name{Yutong Chen$^{a}$, Junhong Zhao $^{\ast b}$  and Wei-Qiang Zhang $^{\ast c}$}

\address{$^{a}$Tsinghua University, chenyt19@tsinghua.org.cn \\$^{b}$Victoria University of Wellington, New Zealand,  junhong.jennifer@gmail.com\\ $^{c}$Tsinghua University, wqzhang@tsinghua.edu.cn.}

\maketitle
\newcommand\blfootnote[1]{%
  \begingroup
  \renewcommand\thefootnote{}\footnote{#1}%
  \addtocounter{footnote}{-1}%
  \endgroup
}

\blfootnote{$^{\ast}$ Junhong Zhao and Wei-Qiang Zhang are corresponding authors.}
\blfootnote{This work was supported by the National Natural Science Foundation of China under Grant No. 62276153.}

\begin{abstract}
    It is in high demand to generate facial animation with high realism, but it remains a challenging task. Existing approaches of speech-driven facial animation can produce satisfactory mouth movement and lip synchronization, but show weakness in dramatic emotional expressions and flexibility in emotion control. This paper presents a novel deep learning-based approach for expressive facial animation generation from speech that can exhibit wide-spectrum facial expressions with controllable emotion type and intensity. We propose an emotion controller module to learn the relationship between the emotion variations (e.g., types and intensity) and the corresponding facial expression parameters. It enables emotion-controllable facial animation, where the target expression can be continuously adjusted as desired. The qualitative and quantitative evaluations show that the animation generated by our method is rich in facial emotional expressiveness while retaining accurate lip movement, outperforming other state-of-the-art methods.
\end{abstract}

%


\section{Introduction}
%
\textit{Facial animation} is a growing research topic that has been widely adopted in many applications, such as education, Virtual Reality (VR), and digital entertainment. Commercial products often require high plausibility and expressiveness in the animated characters to meet users' needs for immersive engagement. Creating such realistic facial animation is a great challenge. 
Speech-driven facial animation is one critical component in this field and has drawn much attention. Our work focuses on speech-driven facial animation in 3D. Compared with animating 2D images, modulating a 3D model is directly applicable to most 3D applications like 3D games and visual aftereffects. Attempts have been made to explore the dependencies between audio and 3D face movement, most of which focus on the lower face and lip movements and their synchronization with speech~\cite{zhou2018visemenet, faceformer,taylor2017deep,tian2019audio2face, cudeiro2019capture,richard2021meshtalk}. Although they can produce plausible basic facial motions, they are far from satisfactory in synthesizing facial expressions, especially in presenting emotions.  

To realize emotional animation based on only an audio track is a challenging task. Although the dependency between sound production and lip movement for the same person is deterministic, its dependencies with the expressions of different emotion categories and intensities are highly ambiguous. There are many different personalized conveying ways of one emotion for different users. Such inherent ambiguity makes neural networks hard to handle emotion variations based on the audio input, even given long-contextual information. The works by Karras et al. and Pham et al.~\cite{karras2017audio,pham2017speech} tried to extract emotion features from speech and implicitly embed them into their neural network to realize emotion synthesis. However, their solution inevitably suffers from over-smoothed regression due to the limited training data, often resulting in limited expressiveness. 

We present a novel approach to realize emotion-controllable facial animation. Instead of recovering lip movement from audio, our method enriches the emotion expressivity and enables the adjustment of the intensity of emotion effects to satisfy the animators' needs. We propose an emotion controller module, which includes an emotion predictor followed by an emotion augment network, to explicitly model the relationship between emotion variations and corresponding facial expression parameters. Image-based emotion recognition was used to generate emotion information as priors to guide the training process.
During inference, the specified emotion condition will apply to the speech-driven facial animation to realize emotion enhancement and customization.
By explicitly modelling emotion, our animator-friendly system enables emotion control with a given emotion type and an intensity value. Our method shows promising results in synthesizing controllable emotional facial animation while retaining high-accuracy lip synchronization, outperforming the state-of-the-arts. \textit{The implementation code is available at \url{https://github.com/on1262/facialanimation}}

\section{Related Work}

Despite much work focusing on facial animation from image or video~\cite{kim2018deep,cao2016real,wei2019vr,thies2016face2face} or generate speech-driven 2D talking head~\cite{wen2020photorealistic,thies2020neural,ji2021audio,yi2020audio,suwajanakorn2017synthesizing}, we concentrate our efforts on speech-driven 3D facial animation, mainly targeting to improve the emotional expression synthesis. We review the most relevant deep learning-based approaches.
\noindent{\textbf{Speech-driven 3D models.}} 
Earlier speech-driven 3D facial animation methods are based on phonetic annotation and viseme-model blending~\cite{zhou2018visemenet,edwards2016jali}. VisemeNet~\cite{zhou2018visemenet} leveraged LSTMs for phoneme grouping and facial landmark prediction, and used their results to regress viseme parameters for lip animation. VOCA~\cite{cudeiro2019capture} tried to encode the identity-dependent information into animation and synthesized various speaking styles. MeshTalk~\cite{richard2021meshtalk} proposed a method to disentangle audio-correlated and audio-uncorrelated information 
to generate more plausible dynamics on the upper face while attaining accurate lip motion. FaceFormer~\cite{faceformer} proposed a transformer-based autoregressive model to encode long-term audio context and synthesize improved lip motions. 
Taylor et al.~\cite{taylor2017deep} proposed a sliding-window regression method to predict the active appearance model (AAM) parameters of a reference lower face based on phoneme labels. Liu et al.~\cite{liu2021geometry} considered the influence of geometry representation and 
utilize them to produce generalized speaker-independent facial animation. Although considerable progress has been achieved in the field, most of these prior works focused on lip motion accuracy without considering expressiveness, which limits the realism of their results.

\noindent{\textbf{Conditional emotion synthesis.}}
Some recent facial animation methods~\cite{ji2021audio,karras2017audio,pham2017speech,chun2021emotion} tackled some issues in emotion synthesis to make more vivid facial animations. Pham et al. used LSTMs in ~\cite{pham2017speech} (improved in~\cite{pham2017end} with CNN-RNN) to model the long-contextual relationship between acoustic features and facial expressions to realize emotion awareness in the generated animation. The method proposed by Karras et al.~\cite{karras2017audio} extracts a latent emotion representation from the audio without identifying emotion categories. In both methods, emotions are implicitly represented and thus lack meaningful guidance to emotion intensity control. Ji et al.~\cite{ji2021audio} 
decomposed speech into emotion and content components to generate emotion-controlled 2D talking heads. 
Chun et al.~\cite{chun2021emotion} introduced an emotion-guided method where emotion-expressive blendshapes are enhanced by emotion recognition guidance and then fused with mouth-expressive blendshapes. Nevertheless, the technique needs manual effort to prepare each emotion template, and the generated expressions often lack upper-face dynamics.

\section{Method}
\subsection{Network Architecture Design}
\label{sec:network_arch}

\begin{figure*}[htbp]
\centering
\includegraphics[width=.90\linewidth]{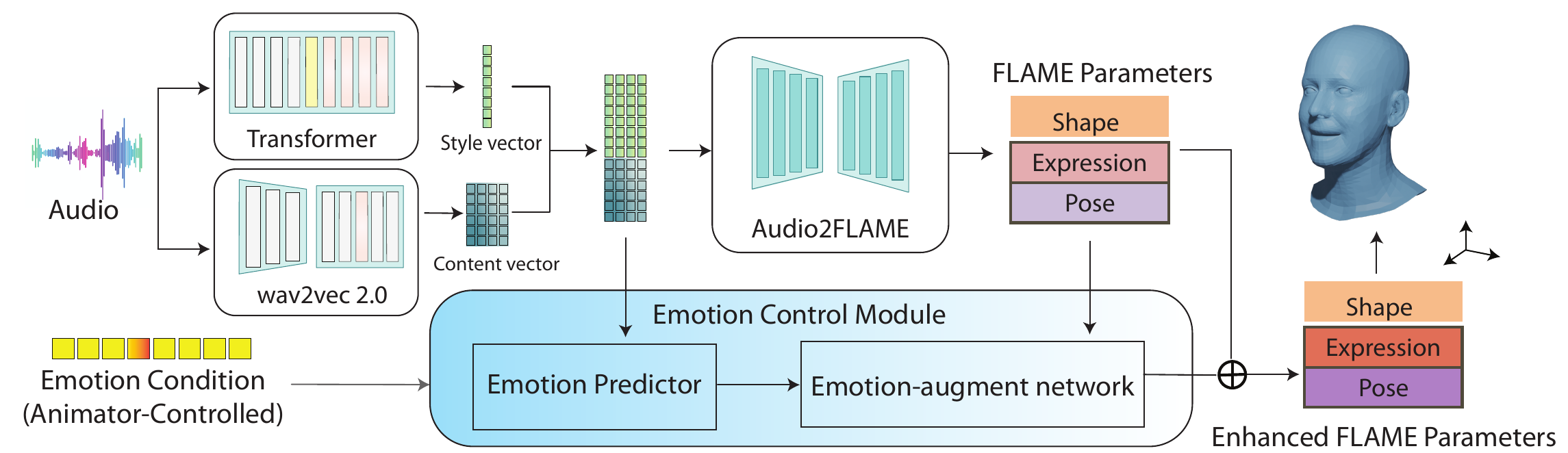}
\vspace{-0.10in}
\caption{An overview of our pipeline.}
\label{fig:pipeline}
\end{figure*}

Figure~\ref{fig:pipeline}. illustrates our pipeline. Our goal is to enrich emotion expressivity in speech-driven facial animation and enable users' control over emotion variations. We first design a neural network to estimate the facial movement represented by FLAME parameters~\cite{li2017learning} (See supplementary) from the input audio, using both local and long-contextual information. The problem can be formulated as follows:
\begin{equation}
\begin{aligned}
(\vec{\phi_{t}} , \vec{\theta_{t}} ) = F(S(\vec{x} _{t\pm \Delta t })\to\vec{\psi_{t}} ; W(\vec{x} _{t\pm \Delta t })\to\vec{\omega_{t}})
\end{aligned}
\end{equation} 
Given the audio segment $x_{t}$ at time $t$ and its neighboring frames, the local content features (content vector $\vec{\omega_{t}}$) and global style features (style vector $\vec{\psi_{t}}$) learned by wav2vec2.0 $W(\cdot)$ and a transformer encoder $S(\cdot)$ are first extracted separately. Then they are concatenated together to predict FLAME parameters. The mapping between the audio and the FLAME parameters is learned by an Auido2FLAME module $F(\cdot)$, which is a multi-layer CNN. The predicted FLAME parameters, including expression parameters $\vec{\phi_{t}}$ and pose parameters $\vec{\theta_{t}}$, combined with the shape parameters of the given identity, are converted to 3D mesh as the output. The lip synchronization will be primarily focused on and preciseness in mouth motion will be ensured at this phase.


We introduce an emotion control module that includes a bi-LSTM-based emotion predictor followed by an embedding layer to generate emotion-related latent features (emotion feature vectors), and a CNN-based emotion-augment network to enhance the expressivity of FLAME parameters based on emotion features. The emotion augmentation process can be represented as:

\begin{equation}
\begin{aligned}
E(\vec{\psi}_{t\pm \Delta t}, \vec{\omega}_{t\pm \Delta t},\gamma_{u,t} ):(\vec{\phi_{t}}, \vec{\theta_{t}})\rightarrow(\vec{\phi^{\prime} _{t}}, \vec{\theta^{\prime}_{t}})
\end{aligned}
\end{equation}
Where $E(\cdot)$ denotes the emotion control module. With the audio features and emotion conditions $\gamma_{u,t}$ customized by the user, it maps $(\vec{\phi_{t}}, \vec{\theta_{t}})$ predicted by the Audio2FLAME model to the emotion-enhanced facial parameters ($\phi^{\prime}_{t}$, $\theta^{\prime}_{t}$). We incorporate the emotion-augment network in a residual manner in our pipeline, which allows dedicated optimization on emotion-related expressions while retaining content-related expressions.
Using this way, the user can explicitly design the emotion intensities and categories at the frame level and regulate the animation output. Additionally, we experimented with other LSTM variants, such as GRU, but found no significant differences in their performance compared to the bi-LSTM model.


\subsection{Emotion Control Module}
\begin{figure}[t!]
\centering
\includegraphics[width=.95\linewidth]{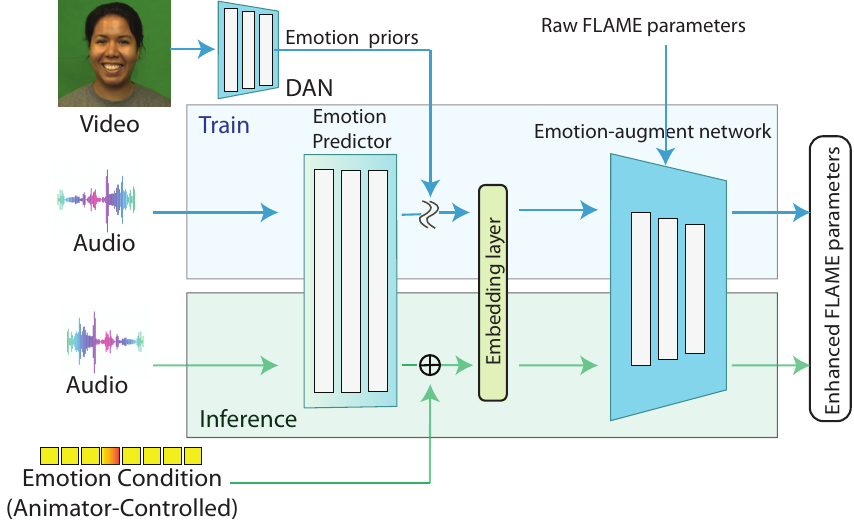}
\vspace{-0.08in}
\caption{Training/inference pipeline of our proposed emotion control module. }
\label{fig:emotion_control}
\end{figure}

The core challenge to realizing full control of emotion simulation is to make the model adaptive to not only emotional state changes, but also emotion strength variations to allow straightforward intensity adjustment. Our proposed training and inference pipeline is illustrated in Fig.~\ref{fig:emotion_control}.
\par
\noindent{\textbf{Emotion prediction and control.}}  In the training phase, to make the network see emotion variations in the input, we leverage the image-based emotion recognition model to obtain frame-level emotion information as emotion priors to facilitate the model training. DAN model from Weng et.al.~\cite{wen2021distract} was used in our experiments, which could be substituted by other similar models. Since emotion in audio is closely related with context, we assume emotion features decoded from 2D visual images are more reliable and informative than that from audio, which could be extracted from a single frame.

However, the emotion classification probabilities from DAN can not indicate the magnitude of emotions. Instead, we found that the emotion logits before the final softmax layer of the emotion recognition network, featuring seven-dimensional vector for seven emotions including happiness, anger and etc., are highly in agreement with the perceived emotion intensity. Therefore, we use them as emotion priors for model training and hinge them with users' emotion control. See supplementary material in detail about our simplified conceptual proof of the linear relation between emotion logits and emotion intensities. In our experiment, we found that the emotion logits are effective in emotion synthesis and work well with the adjustment of both emotion categories and intensities. The emotion augment network is resilient to the prediction error of the emotion priors from the DAN module, and can produce general emotional expressions from them.  

In the inference phase, emotion priors are extracted from audio through a bi-LSTM network, and altered by the customized emotion category and intensity. The bi-LSTM network was trained by maximizing the mutual information between audio-based and video-based emotion priors, using video-based emotion priors as a pseudo ground-truth. The emotion conditions given by the user will be transformed to a one-hot vector with values ranging from 0 to 1 ($\gamma_{u,t}$) and added to the decentralized audio-based emotion priors (segment-level mean normalization ($\overline{\gamma_{a}}$) ), bringing in the final emotion priors $\gamma_{t} = \gamma_{u,t} + (\gamma_{a,t} - \overline{\gamma_{a}} )$.

\begin{figure*}[t!]
\centering
\includegraphics[width=.95\linewidth]{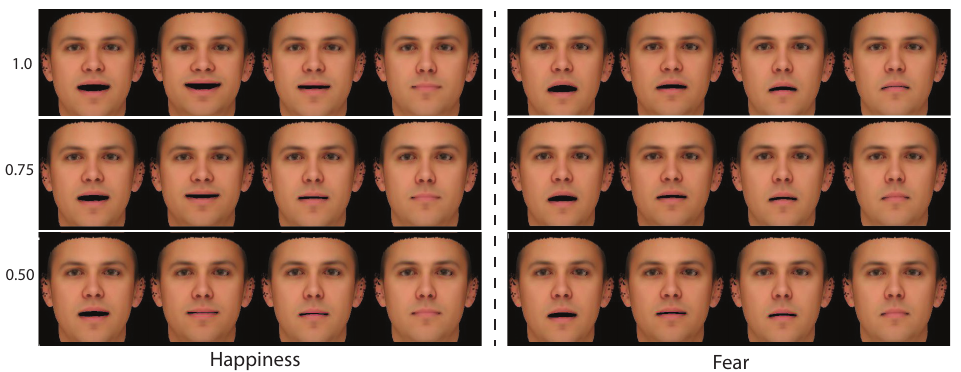}
\vspace{-0.1in}
\caption{Results of emotion animation from the same speech with various emotion classes and customized intensities. The texture is derived from the ground truth video for visualization. Left: animation frames of happiness; Right: animation frames of fear. From top to bottom, the intensities are 1.0, 0.75, and 0.5. }
\label{fig:emoist}
\end{figure*}
\noindent{\textbf{Emotion augment network.}} Before inputting into the emotion augment network, the emotion priors will be transformed into emotion feature vectors through an embedding layer to drive 3D facial expression augmentation. The embedding layer is a learnable 2D matrix that will be optimized during training. The 7-dimensional priors vector will be transformed to a size of 128 emotion feature vectors by multiplying with the learned embedding matrix, and then input into the emotion augment network built with CNN blocks to realize emotion-guided facial expression enhancement. To attain a proper balance between lip synchronization and emotion expressivity, we added the emotion augment network on a residual basis (see~Eq.~\ref{eq:residual}). The raw facial parameters extracted from Audio2FLAME are expected to have a preferable lip synchronization. The gap between the raw facial parameters and ground truth $(\vec{\phi_{gt,t}}; \vec{\theta_{gt,t}})$ is mainly caused by their different emotional expressions, making the augment network $A(\cdot)$ more effective to learn on.
\begin{equation}
A(\vec{\phi_{t}}; \vec{\theta_{t}}) +(\vec{\phi_{t}}; \vec{\theta_{t}})\cong (\vec{\phi_{gt,t}}; \vec{\theta_{gt,t}})
\label{eq:residual}
\end{equation}



\subsection{Loss Design}
We train the model using the loss function:
\begin{equation}
    L = w_{1}L_{vx} +w_{2}L_{lm}
\end{equation}
where $L_{vx}$ is vertex position loss, $L_{lm}$ is mouth shape loss, and $w$ is the weight .

\noindent{\textbf{Vertex position loss:}} A 3D mesh will be transformed from the predicted FLAME parameters, and the L1 difference between the converted vertices and ground truth vertices will be calculated as $L_{vx}$. A vertex mask is applied to the mesh to cover the front face area and exclude ears and eyes. 






\noindent{\textbf{Mouth shape loss:}} 
To ensure lip synchronization, we select the positions of the top ($v_{t}$), bottom ($v_{b}$), leftmost ($v_{l}$), and rightmost ($v_{r}$) vertex and calculate the height of the mouth, $V=|v_{t}-v_{b}|$, and the width of the mouth, $H=|v_{l}-v_{r}|$, as the shape description. The L1 distance of the height and width between ground truth and the predicted mouth shape is used as the mouth shape loss. 




\begin{equation}
    L_{lm} = d_{1} *  ||H_{p}-H_{g}||_{1} + d_{2} * ||V_{p}-V_{g}||_{1}.
\end{equation}
We set $d_{1}=1/0.0476$ and $d_{2}=1/0.017$ in our experiment.

\section{Experiments}
\noindent{\textbf{Data setups:}} Manually-labelled 3D datasets with a rich emotion diversity are limited. Our model is trained using a mixture of 3D and 2D datasets that contain emotion-rich speeches, videos, or 3D model pairs, including VOCASET~\cite{cudeiro2019capture}(3D) and CREMAD\cite{cao2014crema}(2D). For the 2D CREMAD dataset, EMOCA~\cite{danvevcek2022emoca} method was used to reconstruct 3D facial models from 2D images (see EMOCA Results in Supplementary Material) and extract FLAME parameters as the ground truth. 



\subsection{Emotion animation control}
Conditioning on users' control to realize 3D emotional facial animation is one of the key contributions of our method. Figure \ref{fig:emoist} shows controllable facial animation results on happiness and fear emotions with intensities of 0.5, 0.75, and 1.0, respectively. All the examples are driven by the same audio input. The results demonstrate that driven by the same audio input, our model can generate diverse emotional facial animation effects that reveal users' emotion customization. The generated facial expressions were influenced by both users' alteration and audio input. 

Specifically, with the same audio input, the mouth shows a lift in fear rather than a drop in happy emotion. A continuous emotion magnitude change can also lead to satisfied expression variations. For example, in the fear emotion animation, the mouth opens wider, and the mouth corners move downwards with emotion intensity increasing from 0.5 to 1.0. Similar to happiness, in which the mouth corners and cheeks raise harder for the intensity of 1.0 while dropping to neutrally closed for the intensity of 0.5. With our emotion-controllable approach, users can edit the animation in any keyframe by specifying desired intensity values without preparing any other dependencies, which is more straightforward and efficient than traditional methods like~\cite{chun2021emotion}.

\subsection{Comparisons}

We compare our method with the state-of-the-art speech-driven facial animation methods to show how our method performs in both emotion synthesis and lip synchronization. We chose the works by Pham et al.~\cite{pham2017end} and Chun et al.~\cite{chun2021emotion} for the emotion synthesis comparison, and VOCA~\cite{cudeiro2019capture} and FaceFormer~\cite{faceformer} for the lip synchronization comparision (each denoted as [Pham17], [Chun21], [VOCA19] and [FaceFormer22] respectively).

\begin{figure}[t!]
\centering
\includegraphics[width=.95\linewidth]{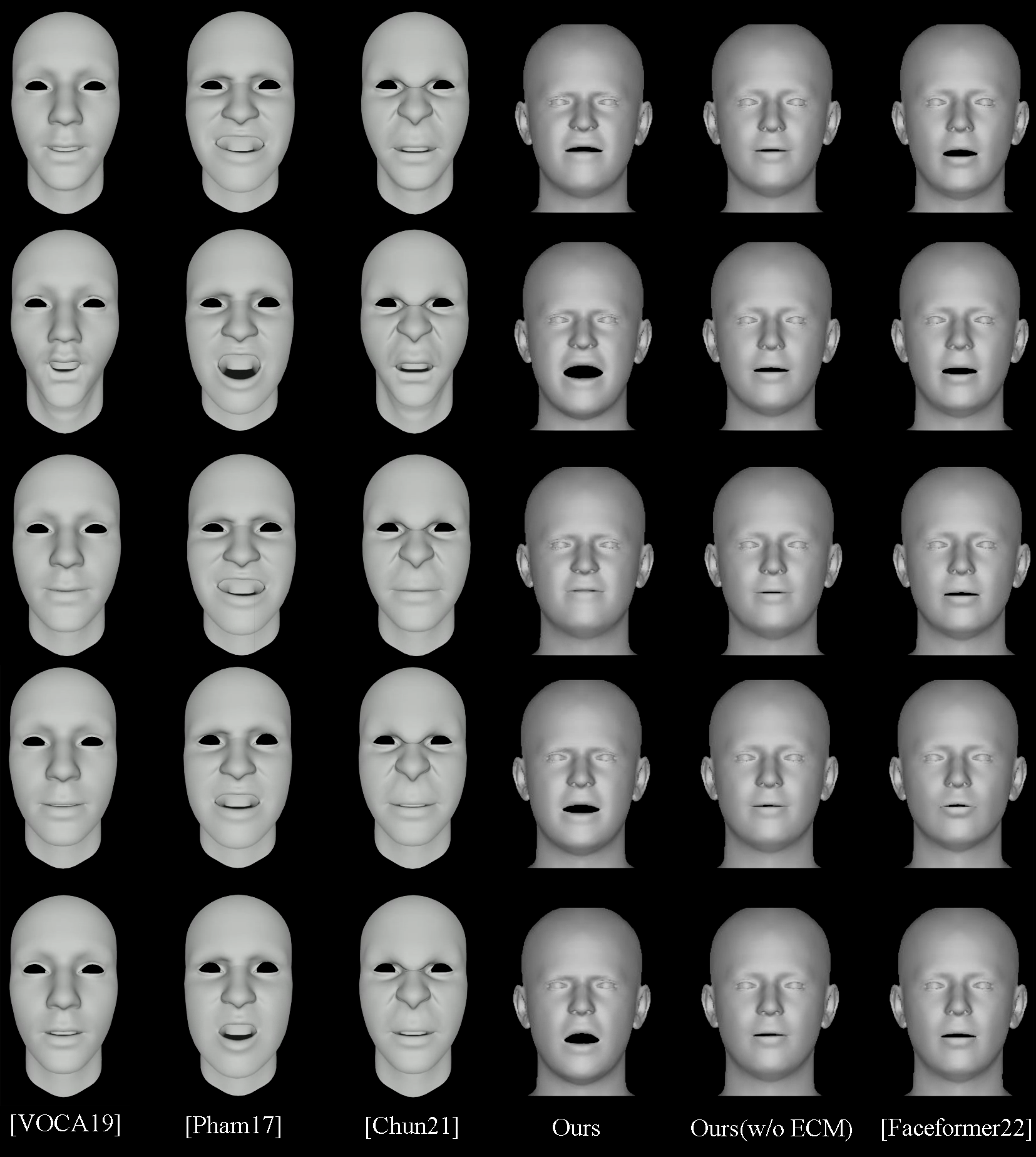}
\caption{Qualitative state-of-the-art comparisons in angry animation. The sentence ``Dogs are sitting by the door" does not come up in our training dataset. Different animation frames were selected from the sentence's beginning (top), intermediate (middle), and end (bottom) phases. \textbf{See the submitted video for more visualizations.}}
\label{fig:emo_cmp}
\end{figure}

\noindent{\textbf{Emotion synthesis.}}
Figure~\ref{fig:emo_cmp} shows an example of facial animation with angry emotion. We observed that [VOCA19] and [FaceFormer22] show accurate lip movement but neutral expressions. While our method can put on an extra layer of emotion variations without sacrificing lip synchronization and comprehensibility of the spoken content. In our results, eyebrows drop, cheek contraction appears, and the degree of mouth opening/closing varies according to mood peaks and troughs. [Pham17] can bring in some emotional expressions, but the lip movement is not as accurate as ours, and the dynamics in the upper face are pretty limited (nearly still). [Chun21] provided emotional expressions with precise lip synchronization, but since the emotion template was given manually and applied uniformly to all frames, the generated animation lacks emotion dynamics alongside the speaking process, especially in the upper face region (see the beginning and end frames). In contrast, ours have more emotional swings along the animation that look natural as real humans expression. In addition, note that our method is capable of transferring emotions from one identity to another, benefiting from the FLAME method's disentanglement of shape and expression, which [Chun21] and [Pham17] are not. The quantitative evaluation of emotion consistency is in supplementary material.

\noindent{\textbf{Lip synchronization.}}
It's essential for speech-driven animation to have accurate lip movements in conveying content information. For a fair comparison in terms of lip synchronization, we used the FLAME parameters from Audio2FLAME to compare with other prior works quantitatively. The testing data keeps the same as what was used by FaceFormer~\cite{faceformer}, which includes two subjects' data from VOCASET, each containing 20 sentence samples. 


To avoid introducing additional alignment errors among different meshes output by different methods, we perform a linear 3D transformation before metrics calculations to make them comparable. After alignment, we uniformly selected 24 key-point vertices around the lip and calculated their distance with ground truth as the measurement of lip synchronization. The lip movement errors of a key-point vertex $k$ in frame $j$ of sequence $i$ are calculated by:
\begin{equation}
    D_{i,j,k}\!=\! \sqrt{(x_{k}\!-\!\hat{x_{k}}) \!+\!(y_{k}\!-\!\hat{y_{k}})\! +\!(z_{k}\!-\!\hat{z_{k}})}
\end{equation}
The overall mean and maximal distance are calculated as two metrics.

Table~\ref{tab:liperr} shows that our method achieved the best results compared with all the other methods, demonstrating that our proposed pipeline is capable of generating decent lip synchronization for emotion enhancement. The RandInit model created from randomized initialization without training steps has the lowest accuracy, as expected. It serves as a reference to show the improvement brought by different training methods. 


\begin{table}[htbp]
\centering
\caption{\bf Comparisons on lip movement error (mm). }
\begin{tabular}{ccccc}
\hline
 &RandInit &[VOCA19] &[FaceFormer22]& Ours \\
 \hline
Mean$\downarrow$ & 2.31 & 1.94 & 1.97 & 1.92 \\
Max$\downarrow$ & 3.87 & 3.41& 3.33& 3.24\\
\hline
\end{tabular}
  \label{tab:liperr}
\end{table}

\subsection{Ablation Study}

\begin{table}[t!]
\centering
\caption{\bf Ablation study measured by lip movement error.}
\vspace{-0.1in}
\begin{tabular}{ccc}
\hline
& \textbf{Mean(mm)$\downarrow$} & \textbf{Maximum(mm)$\downarrow$}\\
\hline
Ours & $\textbf{1.92}$ & $\textbf{3.24}$ \\
\hline
w/o $L_{vx}$ loss & $2.01$ & $3.39$ \\
w/o $L_{lm}$ loss & $2.02$ & $3.34$ \\
\hline
w/o style vector& $2.03$ & $3.38$ \\
\hline
VOCASET-only & $1.97$ & $3.30$ \\
w/ LRS2 & $1.99$ & $3.34$ \\
\hline
\end{tabular}
  \label{tab:ablation}
\end{table}

We ablate the loss items, network components, and training datasets to see their contributions. Table~\ref{tab:ablation} summarizes the effects of our proposed model learned without vertex position loss, without mouth shape loss, and without style vector extracted from the transformer encoder. Our findings demonstrate that both $L_{vx}$ and $L_{lm}$ losses are necessary components of the model's overall loss function, as omitting either lead to decreased performance. Removing any of them caused an increase in error metrics. Style vectors also facilitate the network to capture long-contextual features from the audio input and improve lip synchronization. 

We also tried different training datasets to investigate the benefits of proper training data. Only using the 3D VOCASET dataset cuts down lip movement accuracy. We also investigated the LRS2~\cite{afouras2018deep} dataset. 3000 portrait video clips from BBC television that contain various subjects, background noise, and environments are selected for training. Although the result shows no improvement in lip movement accuracy on the experimental dataset, we observe that it improves our model's generalization on in-the-wild testing.


We also compared the network structure with and without the emotion control module qualitatively. Figure~\ref{fig:emo_cmp} shows the results based on the FLAME parameters from Auido2FLAME (denoted as Ours (w/o ECM)) and the emotion control module (denoted as Ours). We can see that the emotion control module can improve the expressivity of the animation and realize more drastic emotional expression, compared with the neutral animation generated without it.



\section{Conclusion and future work }
We presented a novel deep learning-based approach to generate controllable speech-driven emotional facial animation. An emotion controller module is proposed to enrich emotion expressivity and enables animator customization on emotion intensity and classes. An image-based
emotion recognition was used to generate emotion priors to facilitate explicit emotion learning. Future works could consider pushing the boundary of extreme emotion generation with accurate lip synchronization and improving the animation generation by upgrading the temporal performance of video-based emotion recognition.

{\footnotesize
\bibliographystyle{IEEEbib}
\bibliography{icme_main}
}


\section{Supplementary Materials}

\subsection{FLAME}
\textbf{FLAME} is a statistical 3D head model that is differentiable and disentangles expression and shape. The model includes identity shape parameters $\Vec{\beta}$, pose parameters $\Vec{\theta}$, and facial expression parameters $\Vec{\phi}$. With all the parameters set, FLAME generates a mesh with a set of vertices ($n_{v}=5023$) and faces ($n_{v}=9976$). In FLAME, the shape and expression are disentangled by learning the difference between the neutral and expressive faces of the same person. Some state-of-the-art face reconstruction methods, including DECA and the follow-up EMOCA, used FLAME as the face model and recovered high-detailed and expressive 3D facial expressions from single monocular images. Our work also uses FLAME as the 3D facial model.
\subsection{Emotion Logits as emotion priors in training}



In this work, we use emotional logits generated from the layer before softmax in a facial recognition network as the emotion priors for model training. Here we give a simplified conceptual proof that the emotion logits have a linear relation with the difference in perceptual emotion strength of various emotional states. 
Given a binary (emotion state A and B) classification as an example, we define $a_k$ and $b_k$ as the strength of emotion A and B of sample $S_k$ and $\eta^{a,b}_k=a_k-b_k$ is their difference. 
The possibility of sample $S_k$ to be emotion $A$ is $P(S_k=A|\eta^{a,b}_k)$. We assume that all classes are equally distributed and continuously differentiable. Then, the prior possibility for each class is the same, i.e., $P(S_k=A)=P(S_k=B)$. Therefore, according to Bayes theorem, we get

\begin{equation}
\begin{aligned}
P(S_k\!=\!A|\eta^{a,b}_k)\! = \frac{P(\eta^{a,b}_k|S_k=A)}{P(\eta^{a,b}_k|S_k=A) \!+\!P(\eta^{a,b}_k|S_k=B)}.
\end{aligned}
\end{equation}

We define the joint distribution of ($a_k$, $b_k$) given $S_k=A$ is $f_{a\hat{b}}(\eta)$, and the joint distribution of ($a_k$, $b_k$) given  $S_k=B$ is $f_{\hat{a}b}(\eta)$. And we assume they both follow a normal distribution $N(\mu, \sigma^2)$, then

\begin{equation}
\begin{aligned}
P(\eta^{a,b}_k|S_k=A) = f_{a\hat{b}}(\eta_k) = \int_{C_{1}} f_{a\hat{b}}(a_{k},b_{k})ds
\end{aligned}
\label{eq:fab}
\end{equation}
where $C_{1}$ is $a_k-b_k=\eta_k$. 
With symmetric distribution assumption, we got that:

\begin{equation}
\begin{aligned}
P(\eta^{a,b}_k|S_k\!=\!B)\!=\!\int_{C_{1}}\!f_{\hat{a}b}(a_{k},b_{k})ds\! =\!\int_{C_{2}}\! f_{a\hat{b}}(a_{k},b_{k})ds 
\end{aligned}
\label{eq:fba1}
\end{equation}
where $C_{2}$ is $b_k-a_k=\eta_k$, which can be converted to $a_k-b_k=-\eta_k$. Thus:
\begin{equation}
\begin{aligned}
P(\eta^{a,b}_k|S_k=B) = f_{a\hat{b}}(-\eta_k). 
\end{aligned}
\label{eq:fba2}
\end{equation}        

Considering the softmax projection based on the emotion logits $Z_{ak}$ and $Z_{bk}$ in the emotion recognition network is $P(S_k\!=\!A|\eta^{a,b}_k)\!= \frac{exp(Z_{ak})}{\sum_j exp(Z_{bk})}$, the deviation of emotion logits:

\begin{equation}
\begin{aligned}
\Delta_{z,k}= ln(\frac{P(S_k\!=\!A|\eta^{a,b}_k)}{P(S_k\!=\!B|\eta^{a,b}_k)}) = \ln(\frac{P(\eta^{a,b}_k|S_k=A)}{P(\eta^{a,b}_k|S_k=B)}).
\end{aligned}
\end{equation}






According to formula~\ref{eq:fab},~\ref{eq:fba1}, and~\ref{eq:fba2}, we derive that:
\begin{equation}
    \Delta_{z,k}\!=\!\ln(\frac{f_{a\hat{b}}(\eta_k)}{f_{a\hat{b}}(- \eta_k)})\!=\!\frac{(\eta_k\!+\!\mu)^2\!-\!(\eta_k\!-\!\mu)^2}{2\sigma^2}=\frac{2\mu }{\sigma^2} \eta_k
\end{equation}
which shows the deviation of emotion logits $\Delta_{z,k}$ has linear relation with the deviation of perceptual emotion strength $\eta_k$. The deducing process can be spread to multi-class classification tasks similarly. 

\subsection{Implementation Details}

We train the model end to end with the Adam optimizer, and the learning rate decays from $1e^{-4}$ to $1e^{-5}$ for every ten epochs. 
To effectively model emotions in our pipeline, we extract emotion priors at the frame level to capture the dynamics of emotions over time. However, our image-based prediction method may introduce flickering within an animation sequence. To address this issue, we use a label-smoothing procedure that applies low-pass filtering to the neighboring frames' emotion priors to mitigate temporal inconsistencies. We also refine the predicted FLAME parameters to ensure the bilateral symmetry of the animated face. 

The audio is converted to 16000Hz and is clipped into segments using a sliding window with 100ms windows-length and 33ms stride. Each audio segment and its left and right neighbors ($\Delta_{t}\!=\!1$) are used as the audio input of the network. The video is downsampled to 30 fps and applied with a portrait cropping to adapt them for 3D facial model reconstruction. Wav2vec2 and EMOCA use the model provided by the original paper to extract audio features and reconstruct 3D facial models. Both of them are kept frozen during training. 

For global style vector extraction, we use a transformer encoder that contains four transformer encoder blocks composed of a self-attention layer followed by an addition and normalization layer. The 64-dimension vector output from the last token of the final attention layer is used as the style vector. As a three-layer CNN, Audio2FLAME has a (256, 128, 1, 1, ReLU), (128, 128, 1, 3, ReLU), and (128, 56, 1, 1) setup for input and output channels, kernel size of 1-dimension convolution, stride, and activation functions of each layer. 

The emotion predictor comprises one bi-directional LSTM layer with a 128 hidden size. The predicted emotion priors will be re-scaled with max-min normalization, using minimum and maximum values obtained from statistics of emotion logits extracted from the training video dataset. Emotion-augment network has the same structure as Audio2FLAME, except a layer normalization is applied before the first convolutional layer. With the concatenation of 128-dimensional emotion feature vectors and 56-dimensional raw FLAME parameters from Audio2FLAME as the input (186- dimensions in total), it outputs 56-dimensional enhanced FLAME parameters.

In the training phase, the seed in PyTorch and NumPy randomization is set to 1000. More details can be found in our released code.




\subsection{Emotion synthesis Details}

We use DAN emotion recognition model to assess the resultant emotion expressions generated using our method. The CREMAD testing dataset with 1487 clips was chosen. The output from our model will be rendered into 2D images and then DAN was used to predict the emotions category of them. We randomly select ten frames from each testing video clip. The final prediction result of each clip is the argmax of the predicted categories among the selected ten frames. The confusion matrix is shown in the following Table. ~\ref{table:1}. Our evaluation demonstrated that the majority of emotion categories presented in the generated images were correctly identified by the recognition model, indicating the emotion discrimination of our results. We also observed relatively high levels of ambiguity in some categories in our results, such as between sad and neutral, and fear and anger.

\begin{table}[h!]
\centering
\resizebox{\columnwidth}{!}{\begin{tabular}{||c c c c c c c||} 
 \hline
 Emotions & NEU & HAP & ANG & SAD & DIS & FEA \\ [0.5ex] 
 \hline\hline
 NEU & \textbf{188/217} & 0/217 & 0/217 & 29/217 & 0/217 & 0/217 \\ 
 HAP & 1/254 & \textbf{252/254} & 0/254 & 0/254 & 0/254 & 0/254 \\
 ANG & 4/254 & 0/254 & \textbf{166/254} & 2/254  &  17/254 & \textbf{65/254} \\
 SAD &  \textbf{66/254} & 0/254 & 12/254  & \textbf{176/254} & 0/254 & 0/254  \\
 DIS & 5/254 & 0/254 & 12/254 & 3/254 & \textbf{194/254} & 40/254 \\ 
 FEA & 4/254 & 0/254 & 0/254 & 13/254 & 0/254 & \textbf{237/254} \\[1ex] 
 \hline
\end{tabular}}
\caption{Confusion matrix of DAN recongnition results on our generated animations.}
\label{table:1}
\end{table}

\begin{figure}
\centering
\includegraphics[width=.95\linewidth]{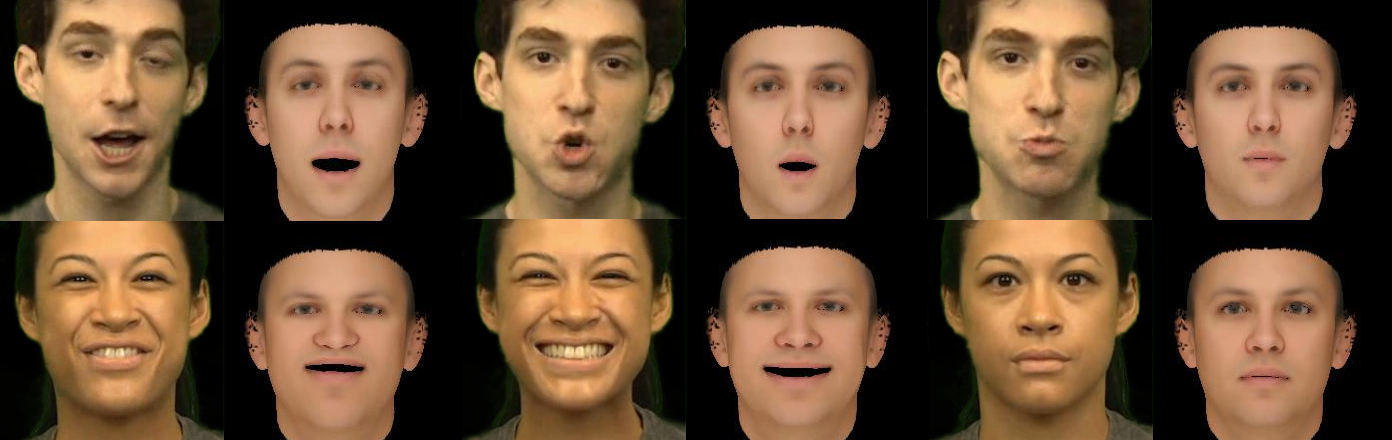}
\caption{Results of EMOCA face geometry reconstruction. }
\label{fig:emoca_result}
\end{figure}

\subsection {EMOCA Results}

Fig. \ref{fig:emoca_result} shows results of EMOCA face geometry reconstruction.

\end{document}